\documentclass[11pt,a4paper]{article}
\usepackage[hyperref]{acl2017}
\usepackage{times}
\usepackage{latexsym}

\usepackage{url}

\aclfinalcopy 

\title{Duluth at SemEval-2017 Task 6:  Language Models in Humor Detection}

\author{Xinru Yan \& Ted Pedersen\\
  Department of Computer Science \\ University of Minnesota Duluth \\ Duluth, MN, 55812 USA \\
  {\tt \{yanxx418,tpederse\}@d.umn.edu}}

\date{}

\begin{document}
\maketitle
\begin{abstract}
This paper describes the Duluth system that participated in 
SemEval-2017 Task 6 \#HashtagWars: 
Learning a Sense of Humor. The system participated in 
Subtasks A and B using N-gram 
language models, ranking highly in the task evaluation. 
This paper discusses the results of 
our system in the development and evaluation stages 
and from two post-evaluation runs. 
\end{abstract}

\section{Introduction}
Humor is an expression of  human uniqueness and intelligence and has drawn 
attention in diverse areas such as linguistics, psychology, philosophy and 
computer science. \textit{Computational humor} draws from all of these 
fields and is a relatively new area of study. 
There is some history of systems that are able to generate 
humor (e.g., \cite{StockS03}, \cite{ozbal2012computational}). 
However, \textit{humor detection} remains a less explored 
and challenging problem (e.g., \cite{Learning:To:Laugh}, 
\cite{Recognizing:Humor:On:Twitter}, \cite{ShahafHM15}, \cite{MillerG15}). 

SemEval-2017 Task 6 \cite{PotashRR17} 
also focuses on \textit{humor detection} by asking participants to 
develop systems that 
learn a sense of humor from the Comedy Central TV show, 
\textit{@midnight with Chris Hardwick}. Our system 
ranks tweets according to how funny they are
by training N-gram language models 
on two different corpora. One consisting of funny tweets 
provided by the task organizers, and 
the other on a freely available research corpus of news data. 
The funny tweet data is made up of tweets that are intended to
be humorous responses to a hashtag given by host Chris Hardwick
during the program. 

\section{Background}

Training \textbf{Language Models} (LMs) is a straightforward 
way to collect a set of rules by 
utilizing the fact that words do not appear in an arbitrary order; 
we in fact can gain useful information about a word by knowing 
the company it keeps ~\cite{Firth57}. A statistical language 
model estimates the probability of a sequence of words or an upcoming 
word. 
An N-gram is a contiguous sequence of N words: a unigram is a 
single word, a bigram is a two-word sequence, 
and a trigram is a three-word sequence. For example, in the tweet 
\begin{quote}
tears in Ramen \#SingleLifeIn3Words
\end{quote}
``tears'', ``in'', ``Ramen'' and ``\#SingleLifeIn3Words'' are unigrams; ``tears in'', 
``in Ramen'' and ``Ramen \#SingleLifeIn3Words'' 
are bigrams and ``tears in Ramen'' and ``in Ramen 
\#SingleLifeIn3Words'' are trigrams.

An N-gram model can predict the next word from a sequence of N-1 previous words.
A trigram Language Model (LM) predicts the conditional probability of the next word using the following approximation:

\begin{equation}
P(w_n|w_1^{n-1})\approx P(w_n|w_{n-2}, w_{n-1})
\end{equation}

The assumption that the probability of a word depends only on a small number of previous words 
is called a \textbf{Markov} assumption ~\cite{markov1954theory}. Given this assumption 
the probability of a sentence can be estimated as follows:
\begin{equation}
P(w_1^n)\approx \prod_{k=1}^{n} P(w_k|w_{k-2}, w_{k-1})
\end{equation}
 
In a study on how phrasing affects memorability, ~\cite{hello} take a language model approach to measure the distinctiveness of
memorable movie quotes. They do this by evaluating a quote with respect to a ``common language'' model built from the newswire sections 
of the Brown corpus ~\cite{BC}. They find that movie quotes which are less like  ``common language'' are more distinctive and therefore
more memorable. The intuition behind our approach is that humor should in some way be memorable or distinct, and so tweets that 
diverge from a ``common language'' model would be expected to be funnier. 

In order to evaluate how funny a tweet is, we train language models on two datasets: 
the tweet data and the news data. 
Tweets that are more probable according to the tweet data language model 
are ranked as being funnier. However, tweets
that have a lower probability according to the news language 
model are considered the funnier since they are the least like the 
(unfunny) news corpus. We relied on both
bigrams and trigrams when training our models. 

We use KenLM ~\cite{Heafield-estimate} as our language modeling tool. Language models are estimated using modified Kneser-Ney smoothing 
without pruning. KenLM also implements a back-off technique so if an N-gram is not found, KenLM applies the lower order N-gram's probability 
along with its back-off weights. 

\section{Method}

Our system\footnote{https://xinru1414.github.io/HumorDetection-SemEval2017-Task6/} estimated tweet probability using N-gram LMs. 
Specifically, it solved the comparison (Subtask A) and semi-ranking (Subtask B) subtasks in four steps:

\begin{enumerate}
\item Corpus preparation and pre-processing: Collected all training 
data into a single file.
Pre-processing included filtering and tokenization.
\item Language model training: Built N-gram language models using KenLM.
\item Tweet scoring: Computed log probability for each tweet based 
on trained N-gram language model. 
\item Tweet prediction: Based on the log probability scores.   

\begin{itemize}
\item Subtask A -- Given two tweets, compare and predict which one is funnier. 
\item Subtask B -- Given a set of tweets associated with one hashtag, rank 
tweets from the funniest to the least funny.

\end{itemize}
\end{enumerate}

\subsection{Corpus Preparation and Pre-processing}

The tweet data was provided by the task organizers. It consists of 106 hashtag files made up of about 21,000 tokens. The hashtag files
were further divided into a development set \textit{trial\_dir} of 6 hashtags and a training set of 100 hashtags \textit{train\_dir}. 
We also obtained 6.2 GB of English news data with about two million tokens from the News 
Commentary Corpus and the News Crawl Corpus from 2008, 2010 and 2011\footnote{http://www.statmt.org/wmt11/featured-translation-task.html}.   
Each tweet and each sentence from the news data is found on a single line in their respective files.

\subsubsection{Preparation}

During the development of our system we trained our language models solely on the 100 hashtag files from \textit{train\_dir}
and then evaluated our performance on the 6 hashtag files found in \textit{trial\_dir}. That data was formatted such that each
tweet was found on a single line.  

\subsubsection{Pre-processing}

Pre-processing consists of two steps: filtering and tokenization. The filtering step was only for the tweet training corpus. 
We experimented with various filtering and tokenziation combinations during the development stage to determine the best setting. 

\begin{itemize}
\item Filtering removes the following elements from the tweets:
URLs, tokens starting with the ``@'' symbol (Twitter user names),
and tokens starting with the ``\#'' symbol (Hashtags). 
\item Tokenization: Text in all training data was split on white space and punctuation
\end{itemize}

\subsection{Language Model Training}

Once we had the corpora ready, we used the KenLM Toolkit to train the N-gram language models on each corpus. 
We trained using both bigrams and trigrams on
the tweet and news data. Our language models accounted for unknown words and were built both with and 
without considering sentence or tweet boundaries. 

\subsection{Tweet Scoring}

After training the N-gram language models, the next step was scoring. 
For each hashtag file that needed 
to be evaluated, the logarithm of the probability 
was assigned to each tweet in the 
hashtag file based on the trained language model.
The larger the probability, the more likely that tweet was according 
to the language model. Table 1 shows an example of two 
scored tweets from hashtag file \textit{Bad\_Job\_In\_5\_Words.tsv} 
based on the tweet data trigram language model.
Note that KenLM reports the log of the probability of the N-grams 
rather than the actual probabilities so the value closer to 0 (-19) has
the higher probability and is 
associated with the tweet judged to be funnier.

\begin{table*}[h!]
\centering
\begin{tabular}{ |p{4.7cm}|p{4.7cm}|p{4.7cm}| } 
\hline
\multicolumn{3}{|c|}{The hashtag: \#BadJobIn5Words} \\
\hline
tweet\_id & tweet & score \\
\hline 
705511149970726912 & The host of Singled Out \#BadJobIn5Words @midnight & -19.923433303833008 \\
\hline
705538894415003648 & Donut receipt maker and sorter  \#BadJobIn5Words @midnight & -27.67446517944336 \\
\hline
\end{tabular}
\caption{Scored tweets according to the trigram LM. The log probability scores computed based on the trigram LM are shown in the third column.}
\label{table:1}
\end{table*}

\begin{table*}[h!]
\centering
\begin{tabular}{ |p{1.2cm}|p{1.2cm}|p{1.2cm}|p{1.7cm}|p{1.5cm}|p{1.9cm}|p{1.7cm}|p{1.7cm}|}
\hline
DataSet & N-gram & \# and @ removed  & Sentence Boundaries & Lowercase & Tokenization & Subtask A Accuracy & Subtask B Distance \\
\hline
\textbf{tweets} & \textbf{trigram} & \textbf{False} & \textbf{False} & \textbf{False} & \textbf{False} & \textbf{0.543} & \textbf{0.887} \\
\hline
tweets & bigram & False & False & False & False & 0.548 & 0.900 \\ 
\hline
tweets & trigram & False & True & True & False & 0.522 & 0.900 \\
\hline
tweets & bigram & False & True & True & False & 0.534 & 0.887 \\
\hline
\textbf{news} & \textbf{trigram} & \textbf{NA} & \textbf{False} & \textbf{False} & \textbf{True} & \textbf{0.539} & \textbf{0.923} \\
\hline
news & bigram & NA & False & False & True & 0.524 & 0.924 \\
\hline
news & trigram & NA & False & False & False & 0.460 & 0.923 \\
\hline
news & bigram & NA & False & False & False & 0.470 & 0.900 \\
\hline
\end{tabular}
\caption{Development results based on \textit{trial\_dir} data. The settings we chose to train LMs are in bold.}
\label{table:2}
\end{table*}

\begin{table*}[h!]
\centering
\begin{tabular}{ |p{1.2cm}|p{1.2cm}|p{1.2cm}|p{1.7cm}|p{1.5cm}|p{1.9cm}|p{1.7cm}|p{1.7cm}|}
\hline
DataSet & N-gram & \# and @ removed  & Sentence Boundaries & Lowercase & Tokenization & Subtask A Accuracy & Subtask B Distance \\
\hline
\textbf{tweets} & \textbf{trigram} & \textbf{False} & \textbf{False} & \textbf{False} & \textbf{False} & \textbf{0.397} & \textbf{0.967} \\
\hline
tweets & bigram & False & False & False & False & 0.406 & 0.944 \\
\hline
\textbf{news} & \textbf{trigram} & \textbf{NA} & \textbf{False} & \textbf{False} & \textbf{True} & \textbf{0.627} & \textbf{0.872} \\
\hline
news & bigram & NA & False & False & True & 0.624 & 0.853 \\
\hline
\end{tabular}
\caption{Evaluation results (bold) and post-evaluation results based on \textit{evaluation\_dir} data. The trigram LM trained on the news data ranked 4th place on Subtask A and 1st place on Subtask B.}
\label{table:3}
\end{table*}

\subsection{Tweet Prediction}

The system sorts all the tweets for each hashtag and orders 
them based on their log probability
score, where the funniest tweet should be listed first. If the scores are based
on the tweet language model then they are sorted in ascending order since the
log probability value closest to 0 indicates the tweet that is most 
like the (funny) tweets model. 
However, if the log probability scores are based on the news data then they
are sorted in descending order since the largest value will have the 
smallest probability associated with it and is therefore least like
the (unfunny) news model.

For Subtask A, the system goes through the sorted list of tweets in a hashtag file
and compares each pair of tweets. For each pair, if the first tweet was funnier 
than the second, the system would output the tweet\_ids for the pair 
followed by a ``1''. If the second tweet is funnier it outputs the tweet\_ids 
followed by a ``0''. For Subtask B, the system outputs all the tweet\_ids for
a hashtag file starting from the funniest. 

\section{Experiments and Results}

In this section we present the results from our development stage (\textit{Table 2}), 
the evaluation stage (\textit{Table 3}), and two post-evaluation results 
(\textit{Table 3}). Since we implemented both bigram and trigam language models during the 
development stage but only results from trigram language models were submitted to the task, 
we evaluated bigram language models in the post-evaluation stage. Note that the accuracy and 
distance measurements listed in Table 2 and Table 3 are defined by the task organizers 
\cite{PotashRR17}. 

Table 2 shows results from the development stage. These results show 
that for the tweet data the best setting is to keep 
the \# and @, omit sentence boundaries, be case sensitive, and ignore 
tokenization. While using these settings the trigram language model
performed better on Subtask B (.887) and the bigram 
language model performed better on Subtask A (.548). We decided to rely
on trigram language models for the task evaluation since the advantage
of bigrams on Subtask A was very slight (.548 versus .543). 
For the news data, we found that the best setting was to 
perform tokenization, omit sentence boundaries, and to be case
sensitive. Given that trigrams performed most effectively in the 
development stage, we decided to use those during the evaluation. 

Table 3 shows the results of our system during the task evaluation. We submitted
two runs, one with a trigram language model trained on the tweet data, and another
with a trigram language model trained on the news data. In addition, after
the evaluation was concluded we also decided to run the bigram language models as well.
Contrary to what we observed in the development data, the bigram language 
model actually performed somewhat better than the trigram
language model. In addition, and also contrary to what we observed with the
development data, the news data proved generally more effective in the
post--evaluation runs than the tweet data. 

\section{Discussion and Future Work}

We relied on bigram and trigram language models because 
tweets are short and concise, and often only consist of just
a few words. 

The performance of our system was not consistent when
comparing the development to the evaluation results.
During development language models trained on the
tweet data performed better.
However during the evaluation and post-evaluation stage, 
language models trained on the news data were 
significantly more effective. We also observed that
bigram language models performed slightly better than
trigram models on the evaluation data. This suggests
that going forward we should also consider both the use of
unigram and character--level language models. 

These results suggest that there are only 
slight differences between bigram and trigram models,
and that the type and quantity of corpora used to train the 
models is what really determines the results. 

The task description paper \cite{PotashRR17} 
reported system by system results for each hashtag. 
We were surprised to find that our 
performance on the hashtag file 
\textit{\#BreakUpIn5Words} in the evaluation 
stage was significantly 
better than any other  system on both Subtask A 
(with accuracy of 0.913) and Subtask B 
(with distance score of 0.636). While we still do not 
fully understand the cause of these results, there is clearly
something about the language used in this hashtag that
is distinct from the other hashtags, and is somehow better
represented or captured by a language model. Reaching a better
understanding of this  result is a high priority for future work. 

The tweet data was significantly smaller than the news data, and
so certainly we believe that this was a factor in the performance
during the evaluation stage, where the models built from the news
data were significantly more effective. Going forward we plan to
collect more tweet data, particularly those that participate in 
\#HashtagWars. We also intend to do some experiments where we 
cut the amount of news data and then build models to see how 
those compare. 

While our language models performed well, there
is some evidence that neural network models
can outperform 
standard back-off N-gram models ~\cite{mikolov2011extensions}. 
We would like to experiment with deep learning methods such 
as recurrent neural networks, since 
these networks are capable of 
forming short term memory and may be better suited for dealing 
with sequence data. 


\end{document}